\documentclass[10pt,twocolumn,letterpaper]{article}

\usepackage{cvpr}              
\usepackage{booktabs}
\usepackage{multirow}
\usepackage[dvipsnames]{xcolor}

\usepackage{colortbl}
\newcommand\hc{ \rowcolor{teal!20}}

\definecolor{cvprblue}{rgb}{0.21,0.49,0.74}
\usepackage[pagebackref,breaklinks,colorlinks,citecolor=cvprblue]{hyperref}

\newcommand{\OURS}{K-Sort Arena }

\def\paperID{*****} 
\def\confName{CVPR}
\def\confYear{2024}

\title{K-Sort Arena: Efficient and Reliable Benchmarking for Generative Models via \\K-wise Human Preferences}

\author{Zhikai Li$^{1,}$\thanks{Equal Contribution}\;, Xuewen Liu$^{1,*}$, Dongrong Joe Fu$^{2}$, Jianquan Li$^{1}$, Qingyi Gu$^{1,}$\thanks{Corresponding Author}\;, Kurt Keutzer$^{2}$, Zhen Dong$^{2,\dag}$ \\
$^1$Institute of Automation, Chinese Academy of Sciences \quad
$^2$University of California, Berkeley\\
{\tt\small lizhikai2020@ia.ac.cn, zhendong@berkeley.edu} \\
{\small \textbf{Project:} \url{https://huggingface.co/spaces/ksort/K-Sort-Arena}}
}

\begin{document}
\maketitle
\begin{abstract}
The rapid advancement of visual generative models necessitates efficient and reliable evaluation methods.
Arena platform, which gathers user votes on model comparisons, can rank models with human preferences.
However, traditional Arena methods, while established, require an excessive number of comparisons for ranking to converge and are vulnerable to preference noise in voting, suggesting the need for better approaches tailored to contemporary evaluation challenges.
In this paper, we introduce K-Sort Arena, an efficient and reliable platform based on a key insight: images and videos possess higher perceptual intuitiveness than texts, enabling rapid evaluation of multiple samples simultaneously. Consequently, K-Sort Arena employs K-wise comparisons, allowing K models to engage in free-for-all competitions, which yield much richer information than pairwise comparisons.
To enhance the robustness of the system, we leverage probabilistic modeling and Bayesian updating techniques.
We propose an exploration-exploitation-based matchmaking strategy to facilitate more informative comparisons. In our experiments, \OURS exhibits 16.3$\times$ faster convergence compared to the widely used ELO algorithm.
To further validate the superiority and obtain a comprehensive leaderboard, we collect human feedback via crowdsourced evaluations of numerous cutting-edge text-to-image and text-to-video models.
Thanks to its high efficiency, \OURS can continuously incorporate emerging models and update the leaderboard with minimal votes. Our project has undergone several months of internal testing and is now available at \href{https://huggingface.co/spaces/ksort/K-Sort-Arena}{K-Sort Arena}.

\end{abstract}

\section{Introduction}
\label{sec:intro}

\begin{figure}[t]
    \centering
    \includegraphics[width=0.8\linewidth]{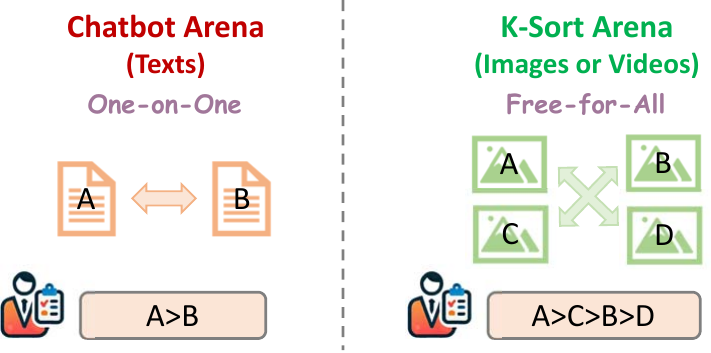}
    \caption{Comparison between K-Sort Arena and Chatbot Arena \cite{chiang2024chatbot}. K-Sort Arena employs K-wise comparisons (K$>$2) to get richer information from user votes. Notably, it introduces probabilistic modeling and an effective matchmaking strategy, significantly improving efficiency and reliability.}
\label{fig:1}
\end{figure}

Visual generation models have made significant advancements, excelling in tasks such as text-to-image \cite{rombach2022sd,podell2023sdxl,betker2023dell3,zhang2023text} and text-to-video \cite{esser2023structure,he2022latent,zhou2022magicvideo,cho2024sora} generation. Such great progress has attracted more and more researchers, leading to a continuous proliferation of new models. Therefore, an efficient and reliable evaluation of the models' capabilities is urgently desired. However, traditional evaluation metrics, such as IS \cite{salimans2016improved}, FID \cite{heusel2017gans}, FVD \cite{unterthiner2018towards}, etc., fall short in providing a fair and comprehensive evaluation, especially not reflecting human preferences in the real world.

\begin{figure*}[t]
    \centering
    \includegraphics[width=0.95\linewidth]{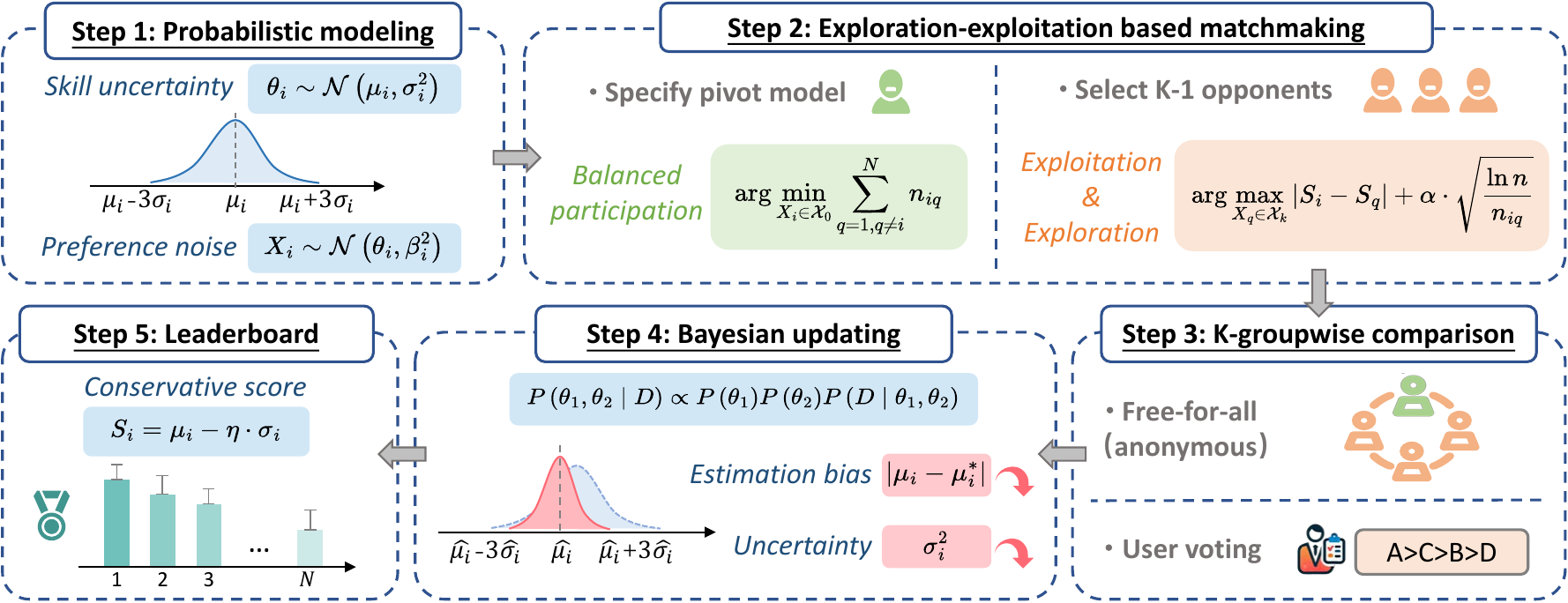}
    \caption{Overview of the proposed K-Sort Arena. K-wise comparisons (K$>$2) and the advanced matching strategy can significantly accelerate ranking convergence, achieving stable ranking with minimal user votes. Probabilistic modeling and Bayesian updating can enhance the robustness of model capability representation, thus resulting in greater efficiency and reliability.}
\label{fig:overview}
\end{figure*}

To this end, Chatbot Arena \cite{chiang2024chatbot} is proposed as a platform developed for evaluating large language models (LLMs). It constructs randomized, anonymous pairwise comparisons of models and collects user judgments of their outputs, thereby forming an overall ranking of models' capabilities. Despite the great progress, Chatbot Arena still faces challenges regarding efficiency and accuracy:
(i) The inefficiency of Chatbot Arena stems primarily from two inherent mechanisms: pairwise comparisons and randomized matching. By allowing only two models to be compared at a time and potentially matching models of vastly different ranks, the system often yields minimal information per comparison. This inefficiency necessitates an excessive number of comparisons to achieve a stable ranking, resulting in a significant waste of valuable human effort in voting. More importantly, as a massive number of new models continuously emerge, this inefficiency prevents the rapid evaluation of new models' capabilities and the timely updating of the leaderboard, causing a lagged response to the latest advances.
(ii) In user voting, preference noise and subjective bias are inherent, leading to occasional unjustified ratings. Pairwise comparisons are sensitive to this issue, which could introduce bias into the relative rankings. This is especially problematic when the leaderboard is updated frequently and the number of votes is small.

To address the above issues, we propose K-Sort Arena, a novel benchmarking platform for visual generation models. K-Sort Arena offers better efficiency and reliability.
Specifically, K-Sort Arena employs K-wise comparisons (K$>$2), allowing K models to participate in free-for-all battles, which provides greater information benefits than pairwise comparisons, as shown in Figure \ref{fig:1}. This approach is based on a practical biological principle: images and videos have higher perceptual intuitiveness compared to texts, enabling rapid evaluation of multiple samples at once. To ensure the robustness of ranking, we introduce probabilistic modeling of the models' capabilities, as well as a Bayesian score updating strategy applied after free-for-all battles among K models, which can dilute the adverse effects of preference noise.
Furthermore, we propose an exploration-exploitation-based model matching strategy, which facilitates matchmaking among models with comparable strength while also incorporating under-explored models, thereby maximizing the expected benefit of each comparison.
The overview is presented in Figure \ref{fig:overview}.

To demonstrate the superiority of \OURS, in Section \ref{sec:experiment}, we design experiments to simulate the scenarios of model comparisons and user voting. Encouragingly, \OURS shows 16.3$\times$ faster ranking convergence than the ELO system in Chatbot Arena and exhibits greater robustness to preference noise. On one hand, it can update the leaderboard accurately and quickly with a minimum number of votes, which can effectively cope with model proliferation; on the other hand, it is more stable and reliable in long-term evaluations, obtaining more trustworthy evaluations with the same number of votes.

K-Sort Arena has served to evaluate dozens of state-of-the-art visual generation models, including both text-to-image and text-to-video models. By statistically organizing user feedback from crowdsourced questions, \OURS effectively builds comprehensive model leaderboards. To aptly reflect the diverse real-world applications, users are free to choose prompts sampled from open-source datasets or to create fresh prompts. 
Moreover, K-Sort Arena supports multiple voting modes and user interactions. Users can either select the best output from a free-for-all comparison or rank the K outputs instead. This flexibility ensures a faster, more user-friendly, and versatile evaluation process. Overall, our contributions can be summarized as follows:
\begin{itemize}
    \item We introduce K-Sort Arena, an efficient and reliable platform for evaluating visual generation models. It can continuously monitor new models and quickly update the leaderboard with minimal votes.
    \item We propose K-wise comparisons to obtain richer feedback information and save human efforts in evaluation.
    \item We devise an exploration-exploitation-based matchmaking strategy with probabilistic capability modeling and Bayesian updating mechanisms.
    \item Ablation study shows that compared to traditional ELO algorithms, \OURS can achieve 16.3$\times$ faster convergence and greater robustness against preference noise.
\end{itemize}
\section{Related Work}
\label{sec:related}

\subsection{Visual Generation Evaluation}
\noindent\textbf{Text-to-Image Benchmarks}
Various metrics have been proposed to assess the performance of text-to-image models \cite{lee2024holistic,zhou2023vision}.
IS \cite{salimans2016improved}, FID \cite{heusel2017gans}, sFID \cite{salimans2016improved}, and KID \cite{binkowski2018demystifying} calculate the distance between the real and generated data distributions using logits or features from InceptionNet \cite{Szegedy2016inception}.
CLIPScore \cite{hessel2021clipscore} computes the cosine similarity between two embeddings from CLIP \cite{radford2021clip}, measuring the alignment of texts and images.
There are also several variants of CLIPScore, such as AS \cite{aesthetic_classifier} and HPS \cite{wu2023human}, which aim to enhance evaluation quality.
Similarly, BLIPScore \cite{bianco2023improving} and ImageReward \cite{xu2024imagereward} are metrics calculated based on BLIP \cite{li2022blip}.
Beyond traditional metrics, recent works introduce large multimodal models as judgment assistants. For instance, T2I-CompBench \cite{huang2023t2i} and X-IQE \cite{chen2023x} utilize the Chain of Thought to enable MiniGPT-4 \cite{zhu2023minigpt} to produce self-consistent evaluations.
VQAScore \cite{lin2024evaluating} employs a visual-question-answering (VQA) model to generate alignment scores by calculating the accuracy of answering simple questions. TIFA \cite{hu2023tifa} also uses VQA to measure the faithfulness of generated images to text inputs.

\noindent\textbf{Text-to-Video Benchmarks}
Metrics for assessing text-to-video models have also been broadly investigated \cite{wu2023cvpr,wu2024towards,liao2024evaluation}.
FVD \cite{unterthiner2018towards} is used to measure the discrepancy between the real and synthesized videos.
CLIPSIM \cite{radford2021learning} is extended to evaluate the alignment of texts and videos by measuring the similarity of multiple frames with texts.
VBench \cite{huang2024vbench} and FETV \cite{liu2024fetv} decompose the evaluation of video quality into multiple dimensions for fine-grained evaluation.
EvalCrafter \cite{liu2024evalcrafter} selects multiple objective metrics, which are expected to summarize real-world situations, to assess the synthesized video quality in various aspects.

Despite the great advances, the above static metrics still suffer significant flaws in expressing human preferences in the real world. They cannot provide comprehensive evaluations, especially in aspects such as visual aesthetics. Furthermore, with the rapid emergence of diverse tasks such as image editing \cite{bai2023integrating}, image captioning \cite{li2022blip}, video editing \cite{ma2024magic}, video captioning \cite{zhou2024streaming}, etc., static metrics are increasingly inadequate in capturing the nuanced performance across these varied and evolving domains.

\subsection{Arena Evaluation with Human Preferences}
To address the limitations of static metrics, DynaBench \cite{kiela2021dynabench} suggests implementing a live benchmark system that integrates a human-in-the-loop approach, thus allowing for more dynamic and adaptive evaluation.
Building on this idea, Chatbot Arena \cite{chiang2024chatbot} is developed as a platform specifically for LLMs. It constructs model arenas that allow LLMs to make randomized, anonymous pairwise comparisons. Users are required to judge and score the outputs of two models to continuously calibrate the capability scores of each model, resulting in an overall ranking of model capabilities. It also inspires WildVision's efforts to rank vision-language models~\cite{lu2024wildvision}.
However, these Arena algorithms require excessive comparisons to achieve a stable ranking and are susceptible to preference noise in voting.
As our concurrent work, GenAI Arena \cite{jiang2024genai} replicates the above workflow to visual generative models and thus has the same issues. Consequently, the coverage of the leaderboard is limited to a few models.
In contrast, K-Sort Arena capitalizes on the intuitive advantage of visual information over texts, incorporating more robust modeling methods and more effective matchmaking strategies, which shows great potential in large-scale model evaluations.
\section{Methodology}
\label{sec:method}

In this section, we describe how to perform robust probabilistic modeling and Bayesian updating of model capabilities in free-for-all comparisons of K models, and how to schedule matches to accelerate ranking convergence.

\subsection{K-wise Comparison}
\label{sec:method_1}

The pairwise comparison employed by Chatbot Arena evaluates only two models per round and is inefficient. In contrast, K-Sort Arena evaluates K models (K$>$2) simultaneously, which naturally provides more information and thus improves the efficiency of the overall ranking. 
In coordination with K-wise comparisons, the modeling and updating of model capabilities are detailed below.

\noindent\textbf{Probabilistic Capability Modeling}
Individual numerical modeling, as in the ELO system \cite{elo1967elo}, provides only a certain value of the estimate and thus cannot ensure reliability. Instead, by using probability distributions to represent capabilities, it is possible to capture and quantify the inherent uncertainty and hence become more flexible and adaptive. This idea can be seen in popular ranking systems such as Glicko \cite{glickman1995glicko} and TrueSkill \cite{herbrich2006trueskill}. Our approach, while inspired by them, incorporates further improvements to enhance efficiency and reliability. Formally, we represent the capability $\theta$ of each model as a normal distribution:
\begin{equation}
    \theta_i \sim \mathcal{N}(\mu_i, \sigma_i^2)
\end{equation}
where $\mu_i$ and $\sigma_i$ denote the $i$-th model's expected score and uncertainty, respectively. Here, $i=1,2,\cdots,N$, and $N$ are the total number of models.
As previously mentioned, user voting inevitably has preference noise, which is orthogonal to the uncertainty $\sigma$ of the model's performance. Therefore, we introduce an additional stochastic variable $\beta$ over the model's capability $\theta$ such that the model's actual performance judged by human evaluation is:
\begin{equation}
    X_i \sim \mathcal{N}(\theta_i, \beta_i^2)
\end{equation}

To build a leaderboard, we use the conservative score \cite{phillips1966conservatism} to estimate the model's capability, as defined below:
\begin{equation}
    S_i^{(n)} = \mu_i^{(n)} - \eta\cdot\sigma_i^{(n)}
\end{equation}
where $\eta$ is a coefficient with a typical value of 3.0, $S_i^{(n)}$, $\mu_i^{(n)}$ and $\sigma_i^{(n)}$ are the values after $n$ comparisons and updates. For each update of $\mu_i^{j}$ and $\sigma_i^{j}$, $j=1,2,\cdots,N$, we follow Eq.~\ref{eq:kwise_update_mu} and Eq.~\ref{eq:kwise_update_sigma} specified as follows.

\noindent\textbf{Bayesian Capability Updating} Based on probabilistic modeling, we implement the updating process using Bayesian inference with observed match results. We begin by discussing the case of two models and then generalize to the free-for-all comparison of K models.
Assuming in the current comparison there are two models $M_1$ and $M_2$, the likelihood estimate of observation $D$ that $M_1$ wins $M_2$ is:
\begin{equation}
P(D|\theta_1, \theta_2) = P\left(X_1>X_2\right)=\Phi\left(\frac{\theta_1-\theta_2}{\sqrt{\beta_1^2+\beta_2^2}}\right)
\end{equation}
Here, $\Phi(x)$ is the cumulative distribution function of standard normal distribution, \ie, $\Phi(x)=\int_{-\infty}^x \phi(u) d u$, and $\phi(x)$ is the probability density function of standard normal distribution, \ie, $\phi(x)=\frac{1}{\sqrt{2 \pi}} e^{-x^2 / 2}$. 
Then, based on Bayes' theorem, we can derive the joint posterior density of ($\theta_1$, $\theta_2$) given observation $D$ as follows:
\begin{equation}
\label{eq:d0}
\begin{aligned}
&P(\theta_1, \theta_2|D) \propto P(\theta_1) P(\theta_2) P(D|\theta_1, \theta_2) \\
&= \phi\left(\frac{\theta_1-\mu_1}{\sigma_1}\right) \phi\left(\frac{\theta_2-\mu_2}{\sigma_2}\right) \Phi\left(\frac{\theta_1-\theta_2}{\sqrt{\beta_1^2+\beta_2^2}}\right)
\end{aligned}
\end{equation}

The marginal posterior density of $\theta_1$ can be subsequently obtained by the following equation:
\begin{equation}
\label{eq:d1}
\begin{aligned}
P\left(\theta_1|D\right) &=\int_{-\infty}^{\infty} P\left(\theta_1, \theta_2|D\right) d \theta_2 \\
&\propto  \phi\left(\frac{\theta_1-\mu_1}{\sigma_1}\right) \Phi\left(\frac{\theta_1-\mu_2}{\sqrt{\beta_1^2+\beta_2^2+\sigma_2^2}}\right)
\end{aligned}
\end{equation}

The derivation of the above equation is detailed in Appendix \ref{app:Bayesian_1}. With the marginal posterior density, the posterior mean of $\theta_1$ can be calculated as follows:
\begin{equation}
\label{eq:d2}
\begin{aligned}
\hat{\mu}_1 &= E\left[\theta_1|D\right] = \frac{\int_{-\infty}^{\infty} \theta_1P(\theta_1|D)d \theta_1}{\int_{-\infty}^{\infty} P(\theta_1|D)d \theta_1} \\
&= \mu_1+\frac{\sigma_1^2}{\sqrt{\sum \left(\beta_i^2+\sigma_i^2\right)}} \frac{\phi\left(\frac{\mu_1-\mu_2}{\sqrt{\sum \left(\beta_i^2+\sigma_i^2\right)}}\right)}{\Phi\left(\frac{\mu_1-\mu_2}{\sqrt{\sum \left(\beta_i^2+\sigma_i^2\right)}}\right)} \\
&= \mu_1+\frac{\sigma_1^2}{c_{12}} \cdot
\mathcal{V}\left(\frac{\mu_1-\mu_2}{c_{12}} \right)
\end{aligned}
\end{equation}
where $\mathcal{V}(x)=\phi (x)/\Phi (x)$ and $c_{ij}^2=\sum\left(\beta_i^2+\sigma_i^2\right)$. The derivation of the above equation is detailed in Appendix \ref{app:Bayesian_2}. Here, $\hat{\mu}_1$ is the updated mean $\mu_1$ value. Similarly, the updating process of the variance $\sigma_1^2$ is given by the following equation:
\begin{equation}
\begin{aligned}
\hat{\sigma_1}^2 &= Var[\theta_1|D] = E[\theta_1^2|D] - (E[\theta_1|D])^2 \\
&= \sigma_1^2\cdot\left(1-\frac{\sigma_1^2}{\sum \left(\beta_i^2+\sigma_i^2\right)} \cdot
\mathcal{W}\left(\frac{\mu_1-\mu_2}{\sqrt{\sum \left(\beta_i^2+\sigma_i^2\right)}} \right)\right) \\
&= \sigma_1^2\cdot\left(1-\frac{\sigma_1^2}{c_{12}^2} \cdot
\mathcal{W}\left(\frac{\mu_1-\mu_2}{c_{12}} \right)\right)
\end{aligned}
\end{equation}
where $\mathcal{W}(x) = \mathcal{V}(x)(\mathcal{V}(x)+x)$.

The above procedure accomplishes Bayesian updating of the two models after comparing them, and as the number of comparisons increases, $\mu$ gets closer to the true value and $\sigma$ tightens up, resulting in a high-confidence capacity estimate \cite{beck1998updating}. We generalize it to a free-for-all comparison of K models, and the capacity updating formulas for the $i$-th model are as follows:
\begin{equation}
\begin{aligned}
\hat{\mu}_i = \mu_i + \sigma_i^2\cdot\Bigg(&\sum_{q:r_q>r_q} \frac{1}{c_{iq}} \cdot \mathcal{V}\left(\frac{\mu_i-\mu_q}{c_{iq}}\right)\\
+& \sum_{q:r_i<r_q} \frac{-1}{c_{iq}} \cdot \mathcal{V}\left(\frac{\mu_q-\mu_i}{c_{iq}}\right)\Bigg)
\label{eq:kwise_update_mu}
\end{aligned}
\end{equation}
\begin{equation}
\begin{aligned}
\hat{\sigma_i}^2 = \sigma_i^2\cdot\Bigg(1-\Bigg( & \sum_{q:r_i>r_q}\frac{\sigma_i^2}{c_{iq}^2} \cdot
\mathcal{W}\left(\frac{\mu_i-\mu_q}{c_{iq}} \right) \\
+& \sum_{q:r_i<r_q}\frac{\sigma_i^2}{c_{iq}^2} \cdot
\mathcal{W}\left(\frac{\mu_q-\mu_i}{c_{iq}} \right) \Bigg) \Bigg)
\label{eq:kwise_update_sigma}
\end{aligned}
\end{equation}

Thanks to the probabilistic modeling and Bayesian updating employed in K-wise comparison, the model's capabilities can be represented with high robustness, thereby facilitating stable and accurate ranking. Additionally, it is important to note that K-wise comparison offers an inherent advantage in terms of efficiency. Generally speaking, a K-wise comparison can be viewed as $C_K^2 = \frac{K(K-1)}{2}$ pairwise comparisons. Assuming that each pairwise comparison provides a certain ranking benefit, and this benefit is additive, we can claim that the total number of comparisons required is significantly less than that for pairwise comparisons.


\subsection{Exploration-Exploitation Based Matchmaking}
\label{sec:method_2}
Effective model matchmaking significantly impacts the efficiency of ranking convergence. Here we first examine matchmaking methods used in notable ranking systems. For instance, the ELO system \cite{elo1967elo}, employed by traditional Arena, uses completely random matching. This can result in pairing the lowest-ranked player with the highest-ranked one, even after numerous comparisons when rankings are nearly stable. Such matchups provide minimal new information, often leading to inefficient use of evaluation resources and slower ranking convergence. 
To address the above issue, TrueSkill system \cite{herbrich2006trueskill} focuses on matching players whose strengths are as equal as possible. However, it is only effective for assessing the ability of individual players, because each player's opponents are limited to a small, localized group of candidates. This limitation means that it lacks a comprehensive understanding of the overall pool of players, making it less useful for the overall ranking of a large number of players.

To this end, we propose an exploration-exploitation-based matchmaking strategy, which promotes valuable comparisons and thus achieves efficient model ranking with fewer comparisons. Specifically, we model the selection of players as a multi-armed bandit problem, where each pair of players is viewed as an arm. The objective is to maximize the overall benefit after $n$ comparisons, \ie, to provide the most information for the overall ranking after $n$ comparisons. Notably, our approach emphasizes maximizing global gains, offering a broader perspective compared to TrueSkill, which focuses on short-term benefits from an individual player's viewpoint.
To solve the multi-armed bandit problem, we apply the Upper Confidence Bound (UCB) algorithm. The UCB algorithm performs exploration with the most optimistic attitude given the current exploitation, which is formulated as follows:
\begin{equation}
    U^{(n)}(X_i, X_q) = |S_i^{(n)}-S_q^{(n)}|+\alpha\cdot\sqrt{\frac{\ln n}{n_{iq}}}
    \label{eq:ucb}
\end{equation}
where $|S_i^{(n)}-S_q^{(n)}|$ indicates the absolute difference in scores between the $i$-th model and $q$-th model after $n$ comparisons, ${n_{iq}}$ denotes the number of comparisons that have been made between the two models, and $\alpha$ is a balancing coefficient with a typical value of 1.0.
Eq. \ref{eq:ucb} realizes the trade-off between exploration and exploitation, where the first part is exploitation and the second part is exploration.
In exploitation phase, we prioritize selecting players of similar skill levels to create valuable comparisons, while in the exploration phase, we encourage players who have not been sufficiently evaluated to participate in matches to ensure a comprehensive assessment.
We theoretically prove its advantage over random selection (details in Appendix \ref{app:ucb}).

Consequently, for a pre-specified player $X_i$, designated as the pivot, we can achieve grouping by greedily selecting its K-1 opponents based on their Upper Confidence Bound (UCB) scores, as follows::
\begin{equation}
\begin{aligned}
\{X_{q_1}^*, \cdots, &X_{q_{K-1}}^*\} = \bigcup_{k=1}^{K-1} \left\{ \arg\max_{X_q \in \mathcal{X}_k } \left\{U(X_i, X_q)\right\} \right\}
\end{aligned}
\end{equation}
\begin{equation}
\begin{aligned}
\text{where} \quad &\mathcal{X}_k = \mathcal{X}_{k-1} - \{X_{q_{k-1}}^*\}, \\
&\mathcal{X}_0=\left\{X_q\right\}_{q=1}^N, \;\; X_{q_0}^* = X_i
\end{aligned}
\end{equation}

The above procedure accomplishes the effective selection of its opponents after specifying the pivot player $X_i$. In our algorithm, instead of random selection, we specify $X_i$ under the guidance of equalizing the number of comparisons to promote balanced participation in comparisons by each player, which is formulated as follows:
\begin{equation}
    X_i = \arg\min_{X_i \in \mathcal{X}_0} \sum_{q=1,q \neq i}^N n_{iq}
    \label{eq:pivot}
\end{equation}

In the following, we present the advantages of the proposed specification policy of the pivot player $X_i$ in a scenario-by-scenario manner.
\begin{itemize}
    \item \textbf{\emph{Scenario 1: Ranking many models from scratch.}} In each round of comparisons, we select the model with the fewest comparisons as the pivot. This promotes balanced participation across all models, preventing insufficient or excessive evaluation of certain models. Such equalization from a global perspective is also an important factor in promoting rapid convergence of the overall ranking.
    \item \textbf{\emph{Scenario 2: Adding new models to an existing ranking.}} Our algorithm facilitates new models to participate in comparisons as pivots frequently in the early rounds so that they can quickly catch up with the number of comparisons of old models. Hence, with an effective matchmaking strategy, we can efficiently evaluate new models' capabilities, allowing us to showcase the latest progress in the leaderboard in real time.
\end{itemize}
\section{Experiments}
\label{sec:experiment}
In this section, we design experiments that simulate user voting and different ranking scenarios, to verify the validity of each proposed component in \OURS.

\noindent\textbf{Experimental Setup}
In Sections \ref{sec:exp_1}, \ref{sec:exp_2}, and \ref{sec:exp_3}, we conduct experiments to rank 50 models from scratch. In Section \ref{sec:exp_4}, we perform experiments by adding a new model to an existing ranking of 50 models.
To simulate user voting on model comparisons, we assign a \emph{preset out-of-order label} to each model to indicate its ground-truth ability. The result of a specific comparison depends on the performance of models, which are determined by their ground-truth abilities and the preference noise. Note that this preset label is used solely for evaluating the comparison results and is not involved in any other part of the ranking process, such as model capability modeling and updating. Finally, we calculate the Mean Squared Error (MSE) of the ranked positions against the preset labels to evaluate the convergence speed and accuracy.

\begin{table}[]\footnotesize
\centering
\caption{Number of comparisons required for ELO system and K-Sort Arena to reach convergence.}
\begin{tabular}{@{}c|ccc|c@{}}
\toprule
Method & K & Modeling & Matchmaking & Comparisons \\ \midrule
ELO System & 2 & Numerical & Random & 11692 \\
\hc K-Sort Arena & 4 & Probabilistic & UCB & 716 ($\downarrow$16.3$\times$) \\ \bottomrule
\end{tabular}
\label{tab:elo}
\end{table}

\subsection{K-Sort Arena vs. ELO-based Arena}
Table \ref{tab:elo} shows the number of comparisons required for ELO system and K-Sort Arena to reach convergence, \ie, MSE becomes consistently zero.
Encouragingly, with the advanced modeling method and matchmaking strategy, K-Sort Arena is 16.3 times more efficient than ELO system, dramatically reducing the number of user votes required. Below we will verify the advantages of each component.

\begin{figure}[b]
    \centering
    \begin{subfigure}{0.85\linewidth}
        \includegraphics[width=1.0\linewidth]{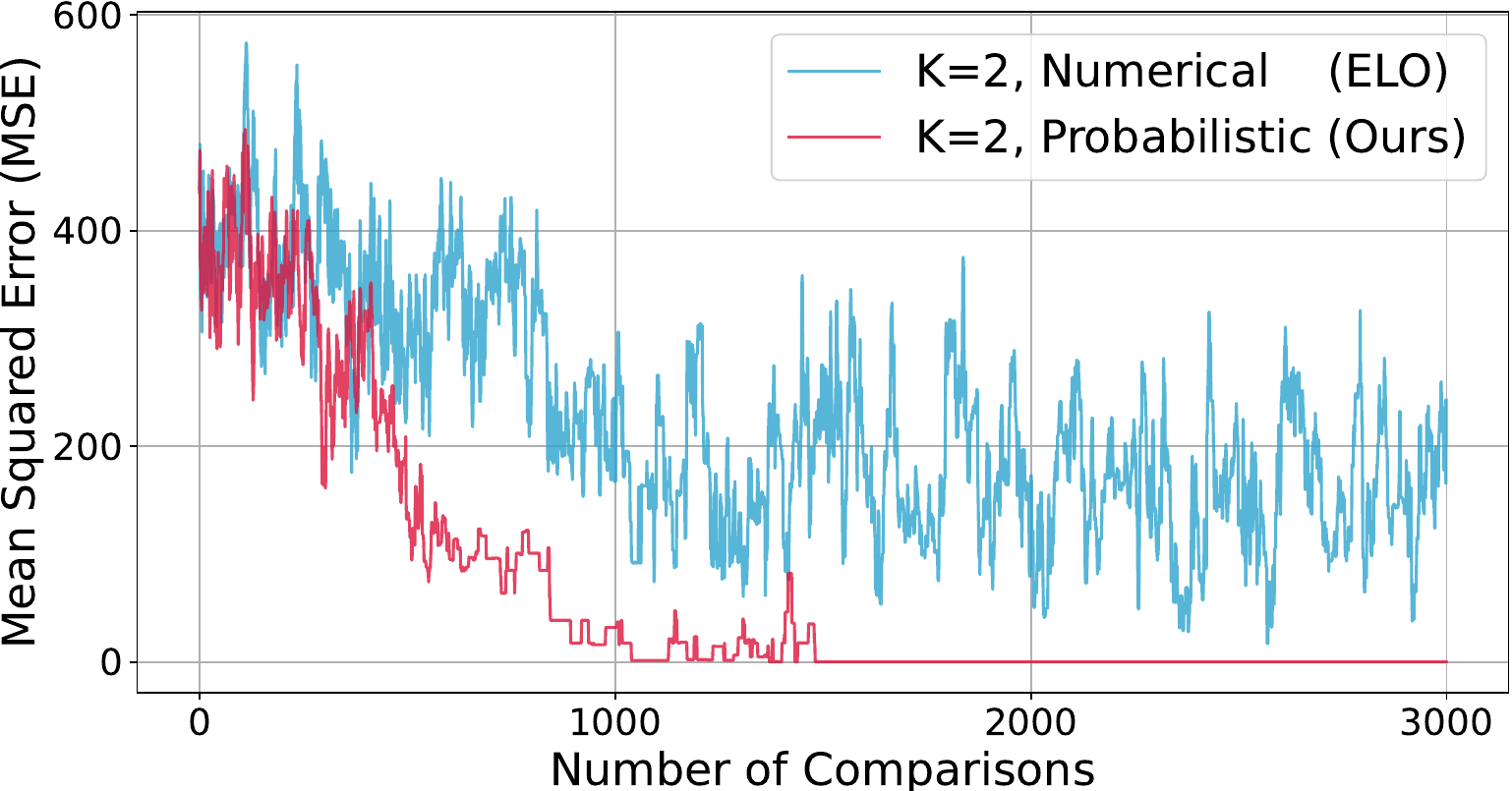}
        \caption{Case without preference noise.}
        \label{fig:exp1-1}
    \end{subfigure} 
    \\
    \begin{subfigure}{0.85\linewidth}
        \includegraphics[width=1.0\linewidth]{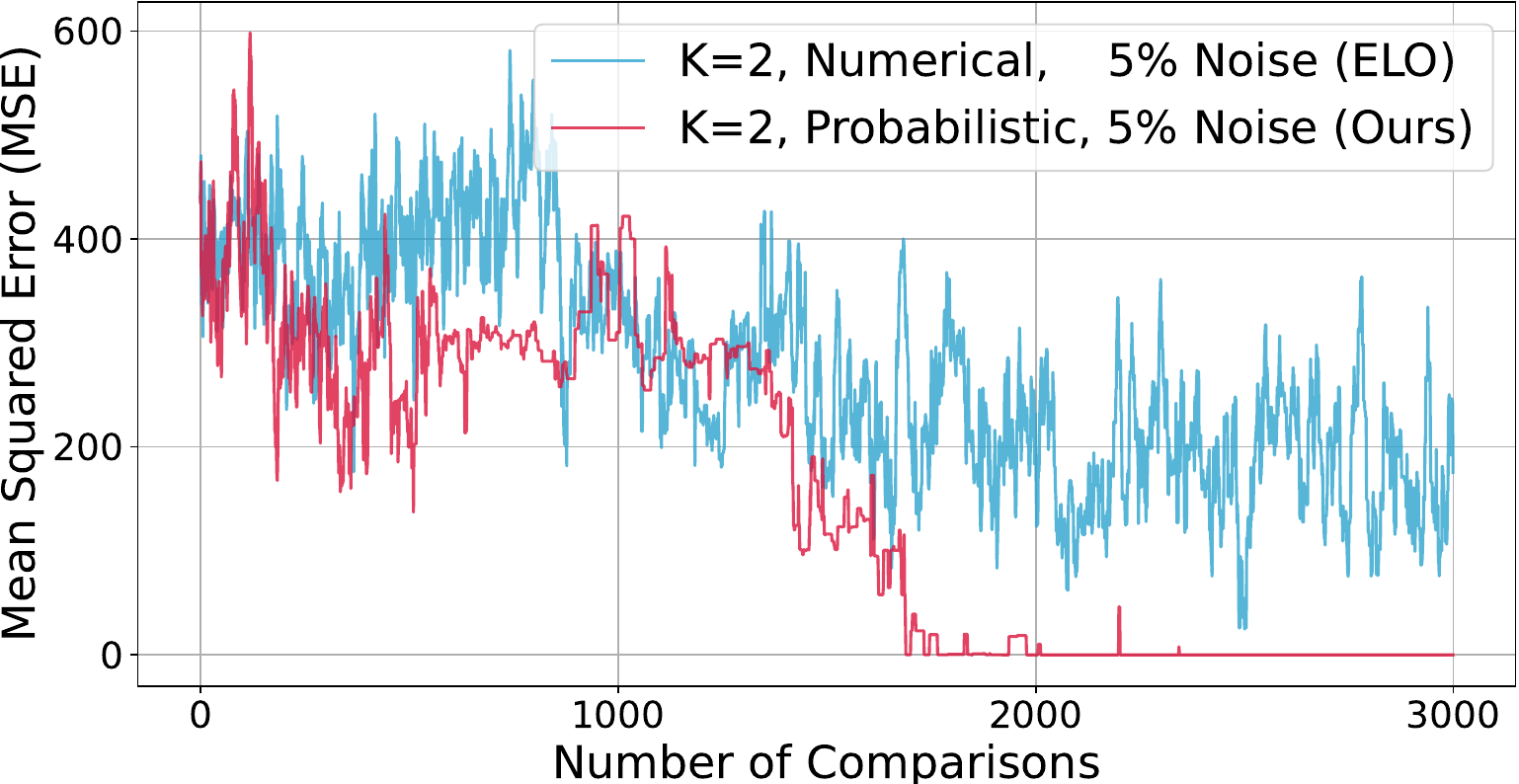}
        \caption{Case with preference noise.}
        \label{fig:exp1-2}
    \end{subfigure}
    \caption{Comparison of numerical modeling (ELO \cite{elo1967elo}) and probabilistic modeling (ours) at K=2, separately with and without preference noise. Probabilistic modeling can converge quickly, while numerical modeling stays oscillating and fails to converge.}
\label{fig:exp1}
\end{figure}

\subsection{Probabilistic vs. Numerical Modeling}
\label{sec:exp_1}
We begin by verifying the advantages of probabilistic modeling over numerical modeling employed in ELO systems, as shown in Figure \ref{fig:exp1-1}.
Since the ELO system is designed for pairwise comparisons, we fix K in \OURS to 2 in this experiment for fairness.
Remarkably, numerical modeling exhibits violent oscillation and fails to converge even after 3000 comparisons. This outcome highlights the unreliability of the existing Arena platform, despite the large number of votes that have been collected. On the contrary, our probabilistic modeling provides rapid convergence after about 1500 comparisons.

Figure \ref{fig:exp1-2} illustrates the case of voting with preference noise. In Figure \ref{fig:exp1-1}, comparison results are directly determined by the preset labels, whereas in Figure \ref{fig:exp1-2} we introduce a 5\% chance of inconsistency between the comparison results and the labels. As observed, numerical modeling still fails to converge, while probabilistic modeling, despite converging slightly slower due to noise effects, manages to converge after approximately 2000 comparisons.
This fully demonstrates the high robustness of probabilistic modeling, offering a strong assurance of the reliability of evaluations.

\subsection{K-wise vs. Pairwise Comparison}
\label{sec:exp_2}

\begin{figure}[h]
    \centering
    \includegraphics[width=0.85\linewidth]{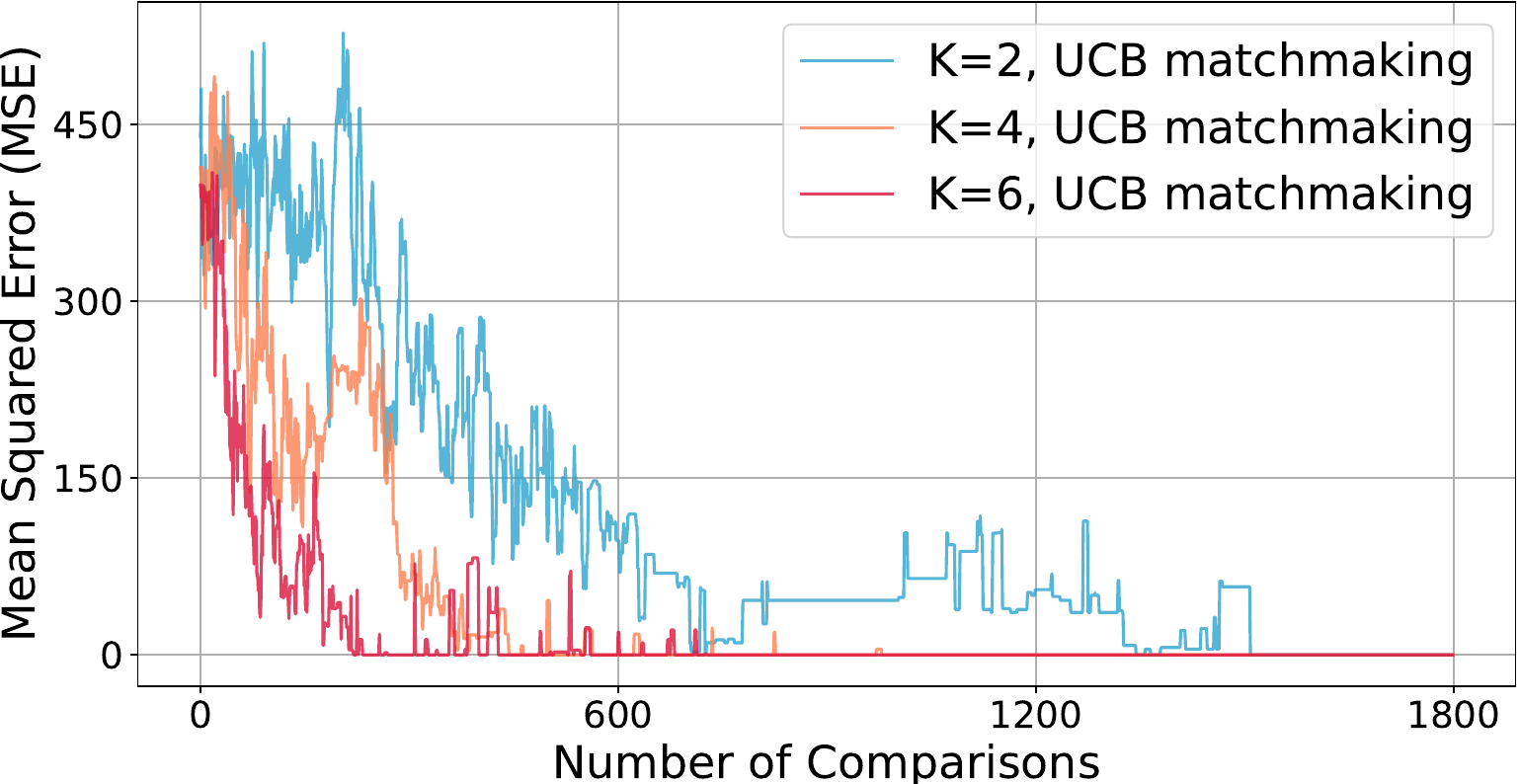}
    \caption{Comparison of different K values when applying UCB matchmaking. K$\in$[2,4,6]. As K increases, the convergence becomes faster and more stable.}
    \label{fig:exp2} 
\end{figure}

Next, we verify the effect of different K values (K$\in$[2,4,6]) on ranking convergence. All three sets of experiments adopt UCB matchmaking strategy, and the experimental results are shown in Figure \ref{fig:exp2}.
When K is increased to 4, multiple models engage in free-for-all comparisons in each round, which yields richer information than the case of K=2, resulting in faster convergence (approximately twice as fast). 
For K=6, while MSE decreases more rapidly in the early stages, small fluctuations occur in the later stages before final convergence, resulting in less pronounced efficiency gains. Therefore, K=4 is considered as a trade-off choice.

\subsection{UCB vs. Traditional Matchmaking}
\label{sec:exp_3}

\begin{figure}[t]
    \centering
    \includegraphics[width=0.85\linewidth]{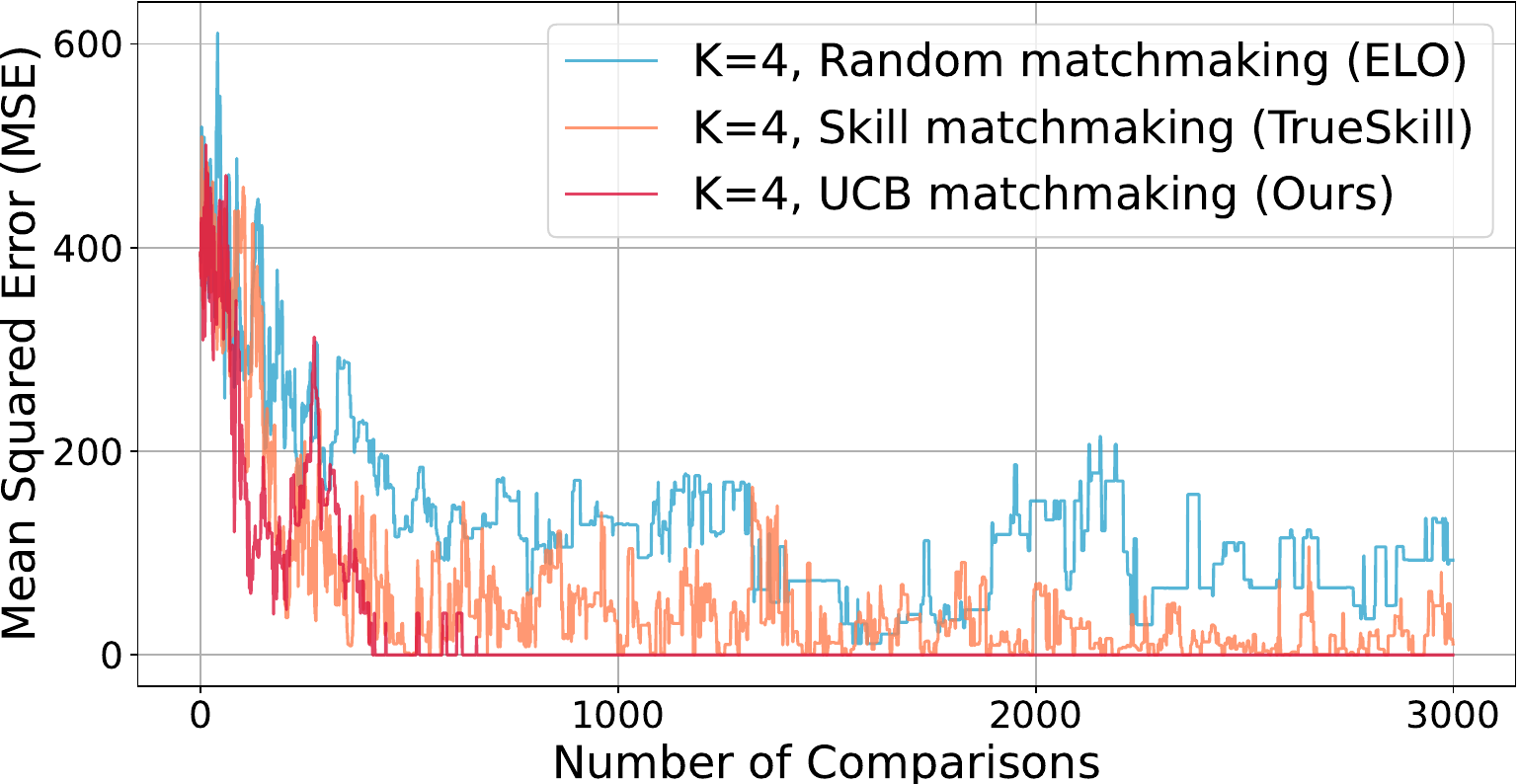}
    \caption{Comparison of different matchmaking strategies at K=4, including random (ELO \cite{elo1967elo}), Skill (TrueSkill \cite{herbrich2006trueskill}), and UCB (ours). The proposed exploitation-exploration based strategy achieves the fastest convergence.}
    \label{fig:exp3} 
\end{figure}

In this section, we demonstrate the advantages of the proposed UCB matchmaking strategy. The comparison methods include random matchmaking in the ELO system and skill-based matchmaking in the TrueSkill system. The experimental results are presented in Figure \ref{fig:exp3}.
Since random matching can potentially result in low-information comparisons, such as pairing the highest-ranked player with the lowest-ranked one, it continues to oscillate after 3,000 comparisons. The goal of skill-based matchmaking is that the skills of players in the comparison are as equal as possible. This may promote interesting matches for an individual player, but it ignores exploration and thus fails to ensure convergence and stability of the overall ranking from a global perspective. Fortunately, our UCB matchmaking strategy addresses this issue by balancing exploitation and exploration, achieving ranking convergence with minimal comparisons.

\begin{figure}[b]
    \centering
    \includegraphics[width=0.85\linewidth]{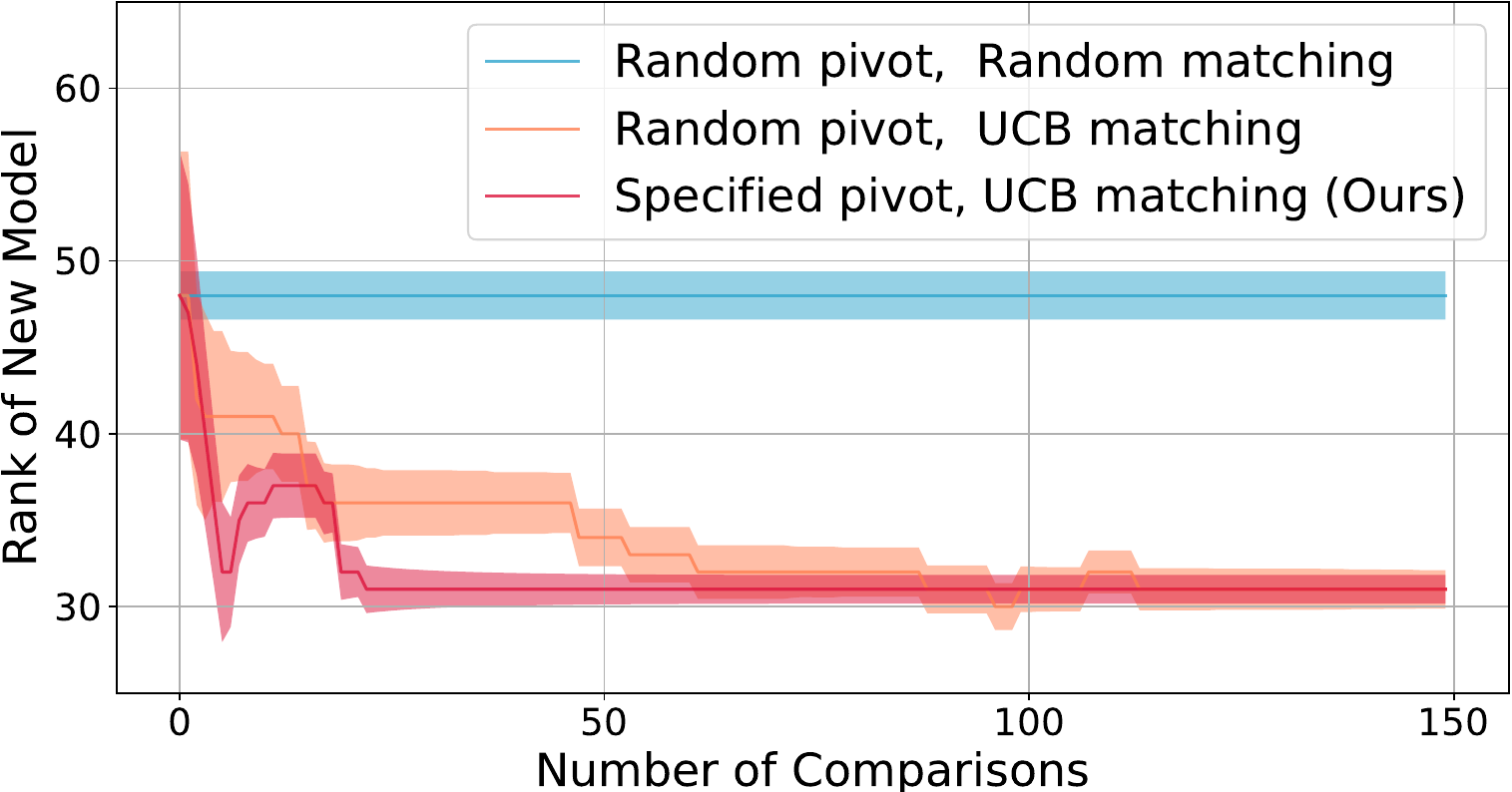}
    \caption{Comparison of different matchmaking strategies when adding a new model. The pivot specification method, coupled with UCB opponent matching, enables the fastest convergence.}
    \label{fig:exp4} 
\end{figure}

\subsection{Specified vs. Random  Pivot}
\label{sec:exp_4}

Here, we focus on the case of adding a new model to an existing ranking and verify the effectiveness of the proposed pivot specification method. We initialize the new model's ranking at 51 and set its actual label to 31. The experimental results are presented in Figure \ref{fig:exp4}.
When both the pivot and its opponents are selected randomly, the new model are less likely to be selected. When the pivot is chosen randomly and UCB is used for matching opponents, the efficiency improves.
This improvement is due to the exploration term in Eq. \ref{eq:ucb}, the new model's small $n_{iq}$ increases its probability of being selected as an opponent. Furthermore, when employing our balance-guided specification method, since the new model naturally participates in the minimal number of comparisons, it is always selected as the pivot in the initial period. Notably, only roughly 30 comparisons are needed to determine the new model's ranking, which provides a prerequisite for rapid leaderboard updating.

\section{K-Sort Arena Platform}
In this section, we build an open and live evaluation platform with human-computer interactions in Huggingface Space, which integrates the proposed algorithms to improve efficiency and reliability. On this platform, users can input a prompt and receive outputs from K anonymous generative models. Users then cast a ranked vote for these models based on their preferred responses, and these votes are saved for updating the leaderboard.
K-Sort Arena platform has the following highlights:
\begin{itemize}
    \item \textbf{\emph{Open-source platform:}} K-Sort Arena platform is open-source, open-access, and non-profit, fostering collaboration and sharing in the community.
    \item \textbf{\emph{Extensive model coverage:}} It covers a comprehensive range of models, including numerous open-source and closed-source models across various types and versions.
    \item \textbf{\emph{Real-time update:}} It continuously adds new models, completes its evaluation with minimal votes, and updates the leaderboard in real-time.
    \item \textbf{\emph{Robust evaluation:}} Bayesian modeling and anonymous comparisons reduce preference noise and model prejudice, making the leaderboard reliable and authoritative.
    \item \textbf{\emph{User-friendly interaction:}} It supports various prompt input modes, voting modes, and user interaction styles, offering users a high degree of flexibility.
\end{itemize}


\subsection{Covered Tasks and Models}
K-Sort Arena is dedicated to evaluating visual generation tasks with human preferences, with a particular focus on text-to-image and text-to-video tasks.
To ensure a comprehensive and thorough evaluation, we strive to cover as many mainstream models as possible, including both open-source and closed-source models, as well as multiple versions of a single model, if available.
Currently, K-Sort Arena has served to evaluate dozens of state-of-the-art models.
A detailed list of models is presented in Appendix \ref{app:list}.

\subsection{Platform Construction}
\OURS platform is designed using Gradio and hosted in Huggface Space. Model inference is performed on ZeroGPU Cloud or Replicate API calls.

\noindent\textbf{Interface Overview}
The interface features two main functionalities: leaderboard display and user voting for model battles. When participating in voting, after the user enters the prompt, the interface can display 4 generated images or videos from the anonymous models, \ie, $K = 4$ is taken as default. 
The interface layout is illustrated in Appendix \ref{app:layout}. 

\noindent\textbf{Prompt Input Mode}
To aptly reflect diverse real-world applications, K-Sort Arena supports two prompt input modes. 
\begin{itemize}
    \item \textbf{\emph{Ready-made prompts:}} Users have the option to randomly extract pre-designed prompts from our extensive data pool for input into the models. This feature eliminates the need for users to spend time creating their own prompts, thereby significantly improving the efficiency of their interactions. At present, the data pool contains 5000 representative prompts, which are sampled from popular datasets such as MS COCO \cite{lin2014microsoft} and WebVid \cite{bain2021frozen}.
    \item \textbf{\emph{Custom prompts:}} Users are also free to create fresh input prompts, allowing them to tailor and customize the generated content to meet their specific needs.
\end{itemize}


\begin{figure}[b]
    \centering
    \includegraphics[width=0.68\linewidth]{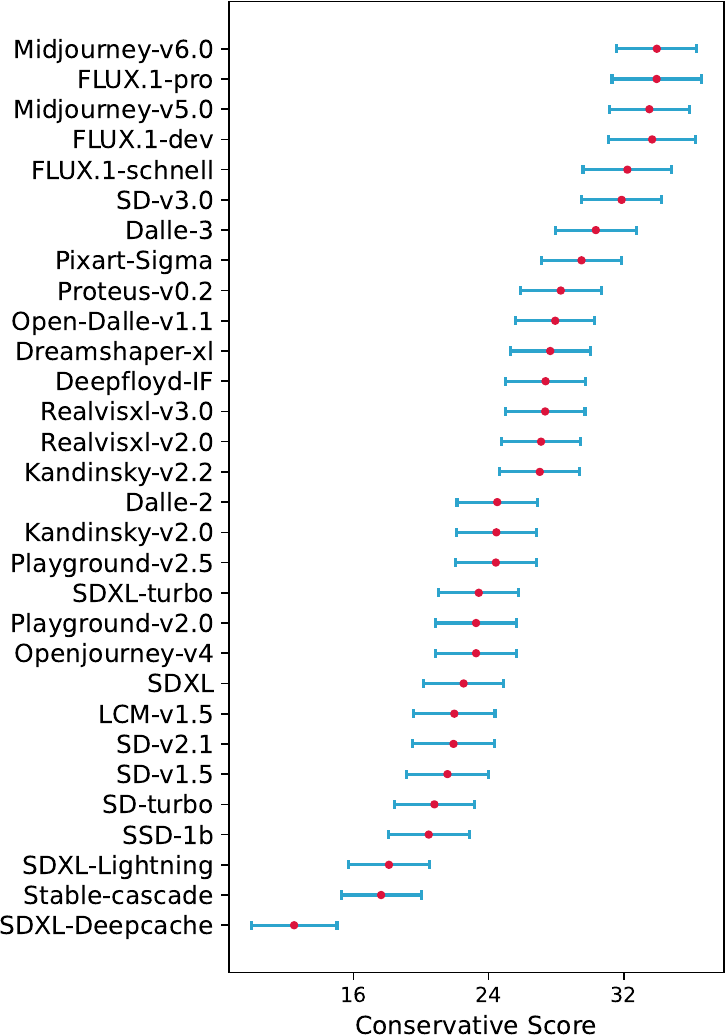}
    \caption{Leaderboard of text-to-image models, which are ranked by conservative score $S$. We also show $\mu$ and $\sigma$ for each model. The data is as of Aug 2024.}
    \label{fig:vote} 
\end{figure}

\noindent\textbf{Voting Mode}
K-Sort Arena supports two voting modes for K-wise free-for-all comparisons, called Best Mode and Rank Mode. 
In Best Mode, users compare the outputs of K models and vote for the most preferred answer. For users who are unsure, a tie option is also available.
In Rank Mode, users can rank the outputs of K models, providing a more fine-grained comparison (tie is also available). 
\begin{itemize}
    \item \textbf{\emph{Best Mode:}} In this mode, the user only needs to select the best model, making one K-wise comparison theoretically equivalent to $K-1$ pairwise comparisons. Since it requires only one mouse click, as in pairwise comparisons, it is $K-1$ times more efficient.
    \item \textbf{\emph{Rank Mode:}} In this mode, the user provides feedback by ranking the K models. One K-wise comparison is theoretically equivalent to $\frac{K(K-1)}{2}$ pairwise comparisons. Since it requires clicking on the rank of each model, \ie, K clicks, it is $\frac{K-1}{2}$ times more efficient.
\end{itemize}



\subsection{Leaderboard Building}
\noindent\textbf{Crowdsourced Voting}
Our project has been undergoing internal testing for several months, during which we have collected over 1,000 high-quality votes. All voters are professors and graduate students in the field of visual generation. To ensure high quality and mitigate preference noise, we organize pre-voting training and provide evaluation guidelines. Specifically, for text-to-image models, the evaluation criteria consist of Alignment (50\%) and Aesthetics (50\%). Alignment encompasses entity, style, and other matching aspects, while Aesthetics includes photorealism, light and shadow rendering, and the absence of artifacts.
Text-to-video models are similarly evaluated based on Alignment (50\%) and Aesthetics (50\%). Alignment is broken down into video content matching, movement matching, and inter-frame consistency. Aesthetics comprises photorealism, physical correctness, and the absence of artifacts.

\begin{figure}[t]
    \centering
    \includegraphics[width=0.68\linewidth]{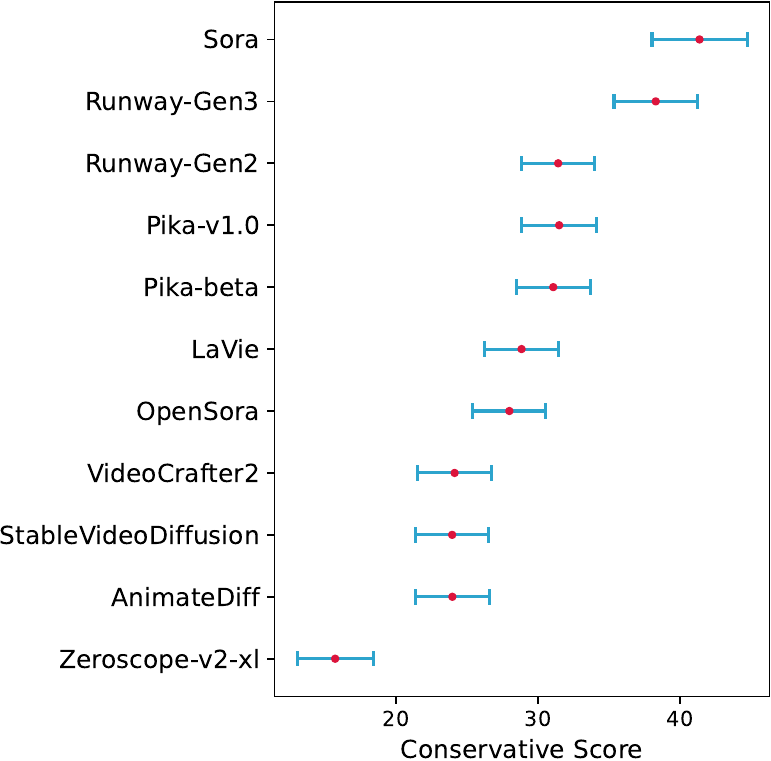}
    \caption{Leaderboard of text-to-video models, which are ranked by conservative score $S$. We also show $\mu$ and $\sigma$ for each model. The data is as of Aug 2024. Note that the comparisons of Sora only take Sora's official samples due to the lack of available API.}
\label{fig:vote_video} 
\end{figure}

\noindent\textbf{Leaderboard Showcase}
The leaderboard of text-to-image models is illustrated in Figure \ref{fig:vote}. We can observe that proprietary models like MidJourney and Dalle dominate the top of the charts. Among open-source models, FLUX.1 and SD-v3.0 stand out with impressive performance. 
The leaderboard of text-to-video models is in Figure \ref{fig:vote_video}.

\label{sec:system}
\vspace{-3mm}
\section{Conclusion}

In this paper, we introduce K-Sort Arena, a benchmarking platform for visual generation models. K-Sort Arena employs K-wise comparisons (K$>$2), allowing K models to play free-for-all games, along with probabilistic modeling and Bayesian updating to improve efficiency and robustness. Furthermore, an exploration-exploitation based matchmaking strategy is proposed to facilitate valuable comparisons, which further accelerates convergence.
We validate the superiority of the proposed algorithms via multiple simulated experiments.
To date, K-Sort Arena has collected extensive high-quality votes to build comprehensive leaderboards for image and video generation. 

\section*{Acknowledgement}
This work was supported in part by the National Science and Technology Major Project under Grant 2022ZD0119402; in part by the National Natural Science Foundation of China under Grant 62276255.

Moreover, we would like to express our gratitude to all those who contributed their time, expertise, and insights during the internal testing phase of K-Sort Arena. Listed in no particular order: Collov Labs; Daquan Zhou from NUS; Yang Zhou from CMU; Vijay Anand from Texas A\&M University; Ying Li, Chun-Kai Fan, Menghang Dong and Aosong Cheng from Peking University HMI Lab; Yinglong from Meituan; Yinsheng Li from Shao's Lab at McGill; Mingfei Guo from Stanford; and Chenyue Cai from Princeton. We are profoundly grateful for their commitment and the unique perspectives they brought to this project.

{
    \small
    \bibliographystyle{ieeenat_fullname}
    \bibliography{main}
}

\clearpage
\setcounter{page}{1}
\maketitlesupplementary

\appendix

\section{Derivation of Bayesian Updating}
\label{app:Bayesian}
In this section, we provide a more detailed derivation of the formulas in Section \ref{sec:method_1} to further clarify the theoretical underpinnings.

\subsection{Derivation of Eq. \ref{eq:d1} in Paper}
\label{app:Bayesian_1}
\begin{equation}
\label{eq:app1}
\begin{aligned}
& P\left(\theta_1|D\right) =\int_{-\infty}^{\infty} P\left(\theta_1, \theta_2|D\right) d \theta_2 \\
& \propto \int_{-\infty}^{\infty} \phi\left(\frac{\theta_1-\mu_1}{\sigma_1}\right) \phi\left(\frac{\theta_2-\mu_2}{\sigma_2}\right) \Phi\left(\frac{\theta_1-\theta_2}{\sqrt{\beta_1^2+\beta_2^2}}\right) d \theta_2 \\
& \propto \phi\left(\frac{\theta_1-\mu_1}{\sigma_1}\right) \int_{-\infty}^{\infty} \phi\left(\frac{\theta_2-\mu_2}{\sigma_2}\right) \Phi\left(\frac{\theta_1-\theta_2}{\sqrt{\beta_1^2+\beta_2^2}}\right) d \theta_2
\end{aligned}
\end{equation}

Now let's focus on the integral part. We first write $\Phi(x)$ as an integral of $\phi(x)$, as follows:
\begin{equation}
\begin{aligned}
    &\Phi \left(\frac{\theta_1-\theta_2}{\sqrt{\beta_1^2+\beta_2^2}}\right) d \theta_2 \\ & = \int_{-\infty}^{\theta_1} \frac{1}{\sqrt{2 \pi (\beta_1^2+\beta_2^2)}} e^{-\frac{\left(y-\theta_2\right)^2}{2\left(\beta_1^2+\beta_2^2\right)}} d y d \theta_2
\end{aligned}
\end{equation}

For simplicity, let $\beta^2=\beta_1^2+\beta_2^2$, and the integral part is as follows:
\begin{equation}
\label{eq:app2}
\begin{aligned}
&\int_{-\infty}^{\infty} \phi\left(\frac{\theta_2-\mu_2}{\sigma_2}\right)
 \int_{-\infty}^{\theta_1} \frac{1}{\sqrt{2 \pi \beta^2}} e^{-\frac{\left(y-\theta_2\right)^2}{2\beta^2}} d y d \theta_2 \\
= & \int_{-\infty}^{\theta_1} \left( \int_{-\infty}^{\infty} \phi\left(\frac{\theta_2-\mu_2}{\sigma_2}\right)
  \frac{1}{\sqrt{2 \pi \beta^2}} e^{-\frac{\left(y-\theta_2\right)^2}{2\beta^2}} d \theta_2 \right) d y \\
= & \int_{-\infty}^{\theta_1} \left( \phi\left(\frac{\theta_2-\mu_2}{\sigma_2}\right) * \phi\left(\frac{y-\theta_2}{\beta}\right) \right) d y \\
= & \int_{-\infty}^{\theta_1} \left( \phi\left(\frac{y-\mu_2}{\sqrt{\sigma^2+\beta^2}}\right)\right) d y \\
= & \Phi\left(\frac{\theta_1-\mu_2}{\sqrt{\beta_1^2+\beta_2^2+\sigma_2^2}}\right)
\end{aligned}
\end{equation}
Where ``$*$'' denotes the convolution of two Gaussian functions. 
Finally, Bringing the above result into Eq. \ref{eq:app1}, we have:
\begin{equation}
\begin{aligned}
P\left(\theta_1|D\right) &=\int_{-\infty}^{\infty} P\left(\theta_1, \theta_2|D\right) d \theta_2 \\
& \propto \phi\left(\frac{\theta_1-\mu_1}{\sigma_1}\right)\Phi\left(\frac{\theta_1-\mu_2}{\sqrt{\beta_1^2+\beta_2^2+\sigma_2^2}}\right)
\end{aligned}
\end{equation}

\subsection{Derivation of Eq. \ref{eq:d2} in Paper}
\label{app:Bayesian_2}

\begin{equation}
\label{eq:app3}
\begin{aligned}
\hat{\mu}_1 &= E\left[\theta_1|D\right] = \frac{\int_{-\infty}^{\infty} \theta_1P(\theta_1|D)d \theta_1}{\int_{-\infty}^{\infty} P(\theta_1|D)d \theta_1} \\
&= \frac{\int_{-\infty}^{\infty} \theta_1 \phi\left(\frac{\theta_1-\mu_1}{\sigma_1}\right) \Phi\left(\frac{\theta_1-\mu_2}{\sqrt{\beta_1^2+\beta_2^2+\sigma_2^2}}\right) d \theta_1}{\int_{-\infty}^{\infty} \phi\left(\frac{\theta_1-\mu_1}{\sigma_1}\right) \Phi\left(\frac{\theta_1-\mu_2}{\sqrt{\beta_1^2+\beta_2^2+\sigma_2^2}}\right) d \theta_1}
\end{aligned}
\end{equation}

We begin with the derivation of the numerator of Eq. \ref{eq:app3}. Again, we write $\Phi(x)$ as an integral of $\phi(x)$, as follows:
\begin{equation}
\begin{aligned}
&\Phi\left(\frac{\theta_1-\mu_2}{\sqrt{\beta_1^2+\beta_2^2+\sigma_2^2}}\right) \\
&= \int_{-\infty}^{\theta_1} \frac{1}{\sqrt{2 \pi (\beta_1^2+\beta_2^2+\sigma_2^2)}} e^{-\frac{\left(y-\mu_2\right)^2}{2\left(\beta_1^2+\beta_2^2+\sigma_2^2\right)}} d y
\end{aligned}
\end{equation}

The computation of the integrals is analogous to the procedure described in Eq. \ref{eq:app2}, which requires reordering the integrals and performing the necessary convolutions. Here, we omit the repetitive steps and directly show the final result as follows:
\begin{equation}
    \Phi\left(\frac{\mu_1-\mu_2}{\sqrt{\beta^2+\sigma^2}}\right)\left(\mu_1+\frac{\sigma_1^2}{\sqrt{\beta^2+\sigma^2}} \frac{\phi\left(\frac{\mu_1-\mu_2}{\sqrt{\beta^2+\sigma^2}}\right)}{\Phi\left(\frac{\mu_1-\mu_2}{\sqrt{\beta^2+\sigma^2}}\right)}\right)
\end{equation}
where $\sigma^2=\sigma_1^2+\sigma_2^2$ and $\beta^2=\beta_1^2+\beta_2^2$.
Similarly, the derivation result for the denominator of Eq. \ref{eq:app3} is as follows:
\begin{equation}
    \Phi\left(\frac{\mu_1-\mu_2}{\sqrt{\beta^2+\sigma^2}}\right)
\end{equation}

Thus, bringing the numerator and denominator results into Eq. \ref{eq:app3}, we have the following:
\begin{equation}
\begin{aligned}
\hat{\mu}_1 &= E\left[\theta_1|D\right] \\
&= \frac{\int_{-\infty}^{\infty} \theta_1P(\theta_1|D)d \theta_1}{\int_{-\infty}^{\infty} P(\theta_1|D)d \theta_1} \\
&= \mu_1+\frac{\sigma_1^2}{\sqrt{\sum \left(\beta_i^2+\sigma_i^2\right)}} \frac{\phi\left(\frac{\mu_1-\mu_2}{\sqrt{\sum \left(\beta_i^2+\sigma_i^2\right)}}\right)}{\Phi\left(\frac{\mu_1-\mu_2}{\sqrt{\sum \left(\beta_i^2+\sigma_i^2\right)}}\right)}
\end{aligned}
\end{equation}

\section{Proof of theoretical advantages of UCB}
\label{app:ucb}
The cumulative regret of the UCB policy grows logarithmically with the number of comparisons $n$, $R_n=\mathcal{O}(\log n)$, providing better long-term performance compared to the linear growth of cumulative regret, $R_n=\mathcal{O}(n)$, of the random selection policy.

\noindent\textbf{\emph{Proof:}} For all K$>$1, if policy UCB is run on K machines having arbitrary reward distributions $P_1\cdots P_k$ with support in [0,1], then its expected regret after $n$ plays is bounded by:
\begin{equation}
\label{eq:app_ucb1}
  R_n^{UCB} \leq \left[ 8 \sum_{i: \mu_i < \mu^*} \left( \frac{\ln n}{\Delta_i} \right) \right] + \left( 1 + \frac{\pi^2}{3} \right) \left( \sum_{j=1}^K \Delta_j \right)
\end{equation}
where $\mu_1\cdots\mu_k$ are the expected values of $P_1\cdots P_k$, $\mu^*$ is the maximum expected value, and $\Delta_i=\mu^*-\mu_i$ for suboptimal selections. Please refer to \cite{auer2002finite} for a detailed derivation of the above equation.

When adopting random selection, \ie, choosing an arm uniformly at random at each play, the expected regret after $n$ plays is:
\begin{equation}
\label{eq:app_ucb2}
R_n^{Rand} = n \cdot \left( \mu^* - \frac{1}{K} \sum_{i=1}^{K} \mu_i \right)
\end{equation}

In the $R_n^{UCB}$ bound in Eq. \ref{eq:app_ucb1}, the first component is a logarithmic term, and the second component is a constant term and independent of $n$, thus $R_n^{UCB}$ has a logarithmic growth $\mathcal{O}(\log n)$. In Eq. \ref{eq:app_ucb2}, $R_n^{Rand}$ has a linear growth $\mathcal{O}(n)$. This indicates that UCB can makes better selections over time, thus achieving a significantly lower cumulative regret compared to random selection.

In our K-Sort Arena system, the lower regret of the applied UCB policy indicates that it makes higher-reward player groupings. This yields more ranking benefits in a single comparison, thus allowing the system to converge more quickly with fewer comparisons.

\section{List of Evaluated Models}
\label{app:list}
The lists of text-to-image and text-to-video models covered by K-Sort Arena are shown in Table \ref{tab:list1} and Table \ref{tab:list2}, respectively.
The data is in no particular order.
We will continue to add new models. In the future, besides distilled models~\cite{luo2023latent,sauer2023adversarial}, we also plan to include the evaluation of models that are compressed through quantization \cite{li2023vit,li2023repq,li2022patch,li2023q} and pruning \cite{han2015deep,castells2024ld}.

\begin{table}[t]
\caption{List of text-to-image models in K-Sort Arena (in no particular order). Here, we show the name and license of each model.}
\centering
\footnotesize
\setlength{\tabcolsep}{3pt}
\begin{tabular}{@{}cccc@{}}
\toprule
Task & Model & License & Organization \\ \midrule
\multirow{30}{*}{Text2Image} 
& Dalle-3 &  Commercial   & OpenAI \\
& Dalle-2 &  Commercial    &  OpenAI \\
& Midjourney-v6.0 &  Commercial    &   Midjourney  \\
& Midjourney-v5.0 &  Commercial  &  Midjourney  \\
& FLUX.1-pro &  Open source  & Black Forest Labs  \\
& FLUX.1-dev &  Open source  & Black Forest Labs  \\
& FLUX.1-schnell &  Open source  & Black Forest Labs  \\
& SD-v3.0 &  Open source  & Stability AI  \\
& SD-v2.1 &  Open source  &  Stability AI \\
& SD-v1.5 &  Open source  &  Stability AI \\
& SD-turbo &  Open source  &  Stability AI \\
& SDXL &  Open source   & Stability AI \\
& SDXL-turbo &  Open source   & Stability AI \\
& Stable-cascade &  Open source  &  Stability AI \\
& SDXL-Lightning &  Open source  &  ByteDance \\
& SDXL-Deepcache &  Open source  &  NUS \\
& Kandinsky-v2.2 &  Open source  &  AI-Forever \\
& Kandinsky-v2.0 &  Open source  &  AI-Forever \\
& Proteus-v0.2 &  Open source   & DataAutoGPT3 \\
& Playground-v2.5 &  Open source  &  Playground AI  \\
& Playground-v2.0 &  Open source  &  Playground AI \\
& Dreamshaper-xl &  Open source   &  Lykon \\
& Openjourney-v4 &  Open source   &  Prompthero\\
& LCM-v1.5 &  Open source   & Tsinghua \\
& Realvisxl-v3.0 &  Open source   & Realistic Vision \\
& Realvisxl-v2.0 &  Open source   & Realistic Vision \\
& Pixart-Sigma &  Open source   & PixArt-Alpha \\
& SSD-1b &  Open source  &  Segmind \\
& Open-Dalle-v1.1 &  Open source   & DataAutoGPT3 \\
& Deepfloyd-IF &  Open source   & DeepFloyd \\
\bottomrule
\end{tabular}
\label{tab:list1}
\end{table}

\begin{table}[t]
\caption{List of text-to-video models in K-Sort Arena (in no particular order). Here, we show the name and license of each model.}
\centering
\footnotesize
\setlength{\tabcolsep}{3pt}
\begin{tabular}{@{}cccc@{}}
\toprule
Task & Model & License & Organization \\ \midrule
\multirow{11}{*}{Text2Video} 
& Sora &  Commercial   & OpenAI \\
& Runway-Gen3 &  Commercial   & Runway \\
& Runway-Gen2 &  Commercial   & Runway \\
& Pika-v1.0 &  Commercial    &  Pika \\
& Pika-beta &  Commercial    &   Pika  \\
& OpenSora &  Open source   & HPC-AI \\
& VideoCrafter2 &  Open source   & Tencent \\
& StableVideoDiffusion &  Open source   & Stability AI \\
& Zeroscope-v2-xl &  Open source  &  Cerspense \\
& LaVie &  Open source   & Shanghai AI Lab \\
& Animate-Diff &  Open source   & CUHK etc. \\
\bottomrule
\end{tabular}
\label{tab:list2}
\end{table}



\section{Analysis of Votes}
\label{app:votes}
\begin{figure}[t]
    \centering
    \includegraphics[width=1.0\linewidth]{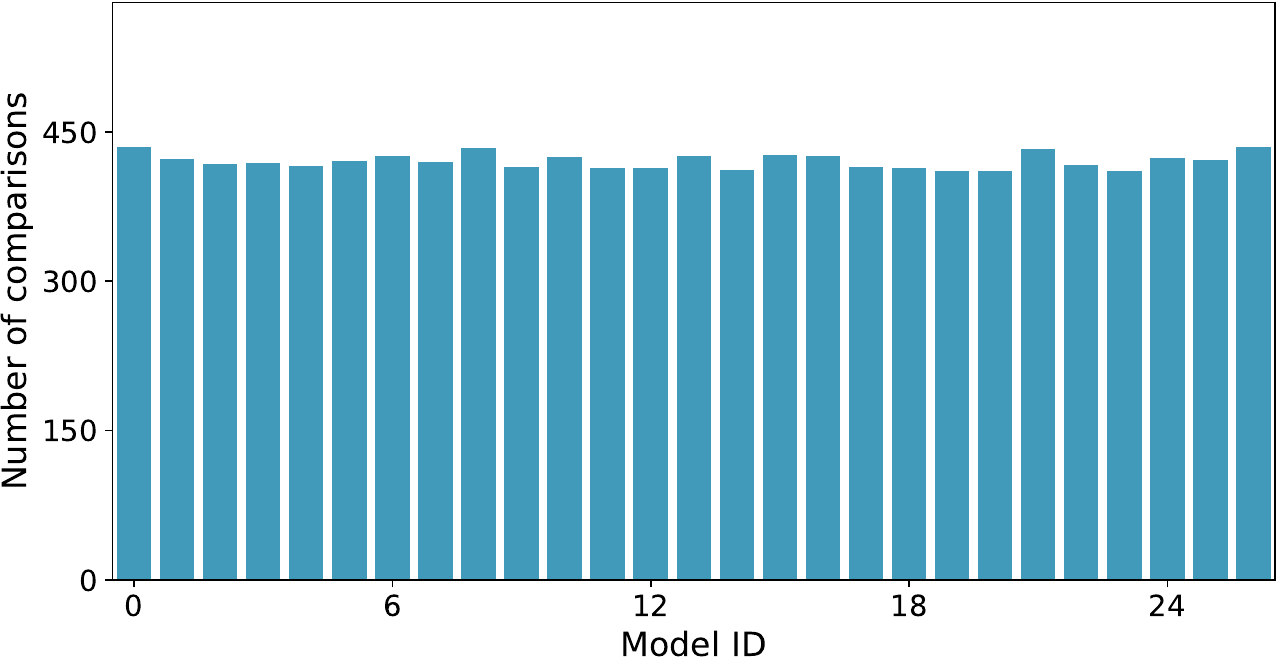}
    \caption{The number of comparisons in which each model is involved. Model IDs are aligned with the order in Table \ref{tab:list1}. The data is as of Aug 2024.}
    \label{fig:vote_1} 
\end{figure}

After several months of internal testing, we have collected over 1,000 votes from experts in the field of visual generation.
Note that in each vote, four models participate in a free-for-all comparison, which is equivalent to $\frac{K(K-1)}{2}=6$ pairwise comparisons. This means our voting process can be approximately converted to over 6,000 pairwise comparisons.
Figure \ref{fig:vote_1} illustrates the number of comparisons in which each model is involved, with the data representing the number of pairwise comparisons after conversion. Thanks to the UCB algorithm and the pivot specification strategy, all models are fully and balanced evaluated.

\begin{figure}[t]
    \centering
    \includegraphics[width=1.0\linewidth]{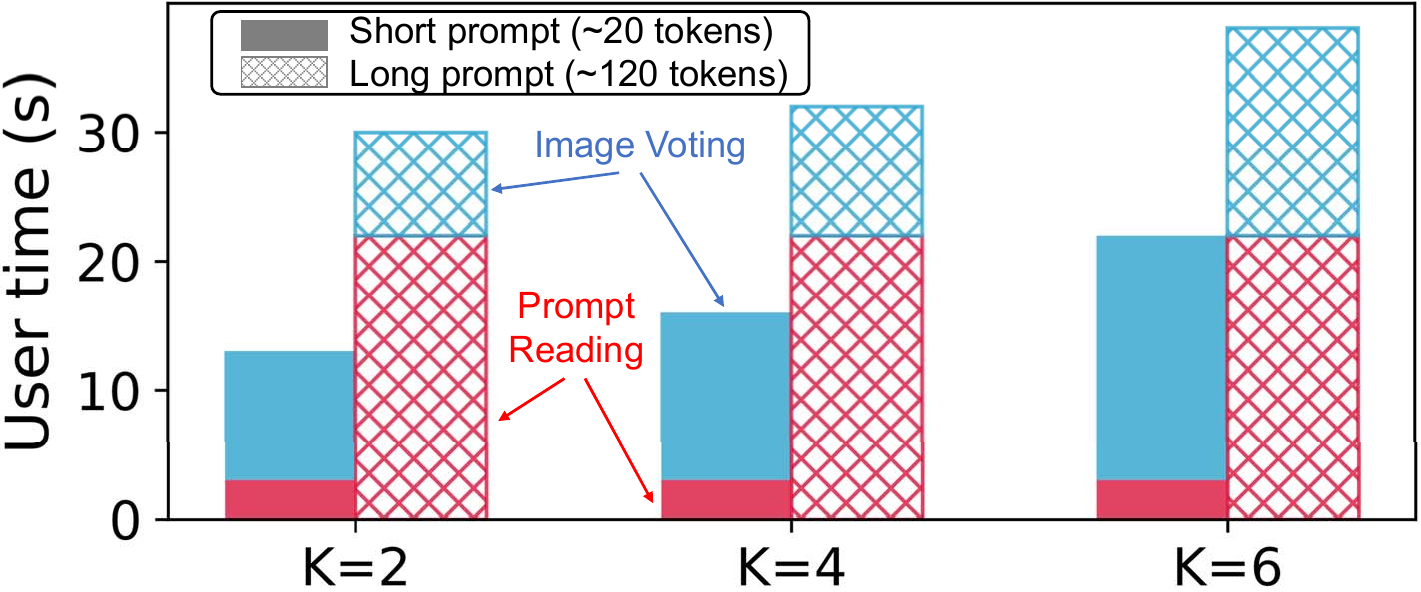}
    \caption{Analysis of user voting times with different K values (2, 4, 6) and prompt complexities.}
    \label{fig:rebuttal_1} 
\end{figure}

\begin{figure*}[t]
    \centering
    \begin{subfigure}{0.90\linewidth}
        \includegraphics[width=1.0\linewidth]{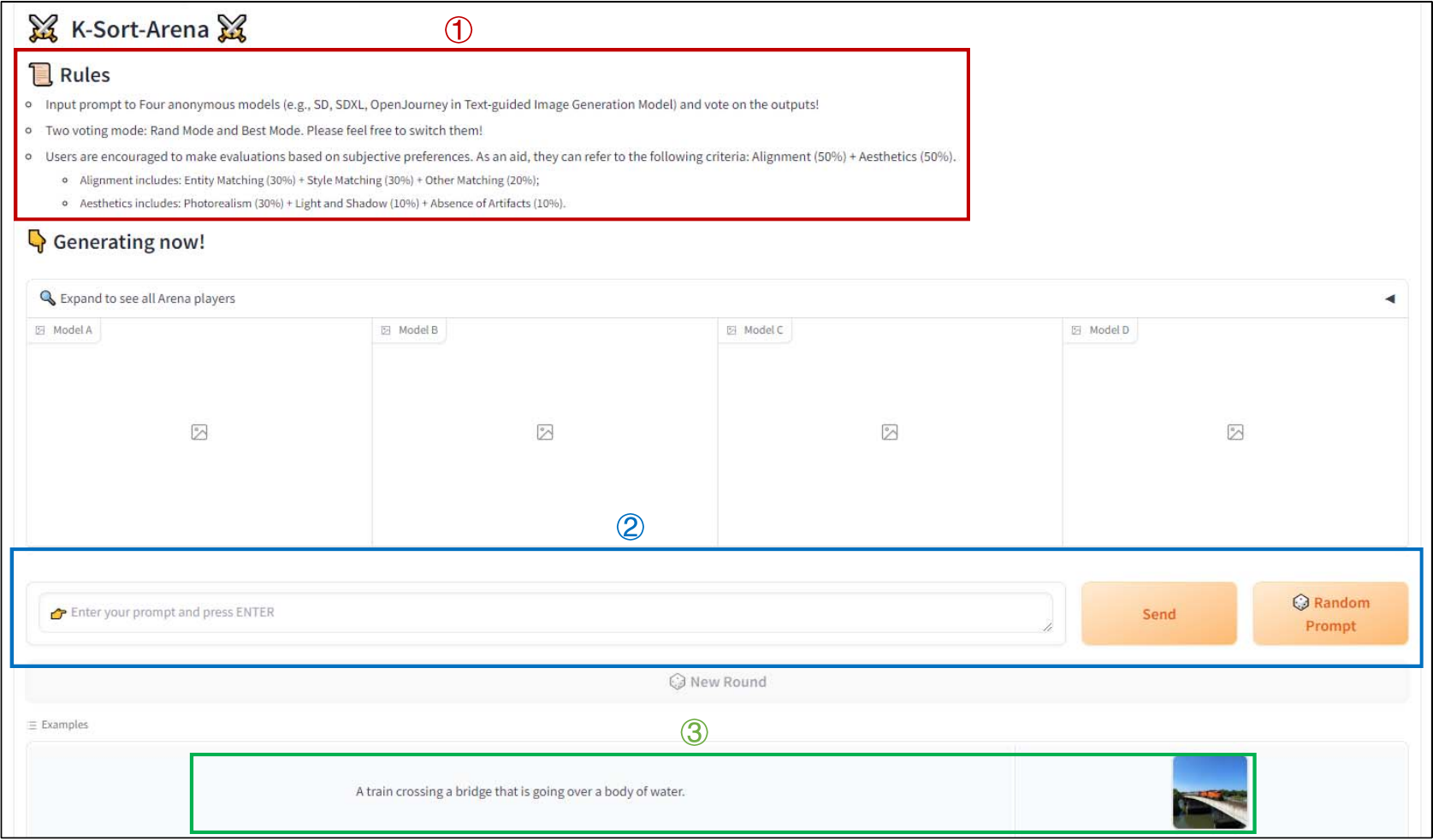}
        \caption{Display of the initial interface.}
    \end{subfigure} 
    \\
    \begin{subfigure}{0.91\linewidth}
        \includegraphics[width=1.0\linewidth]{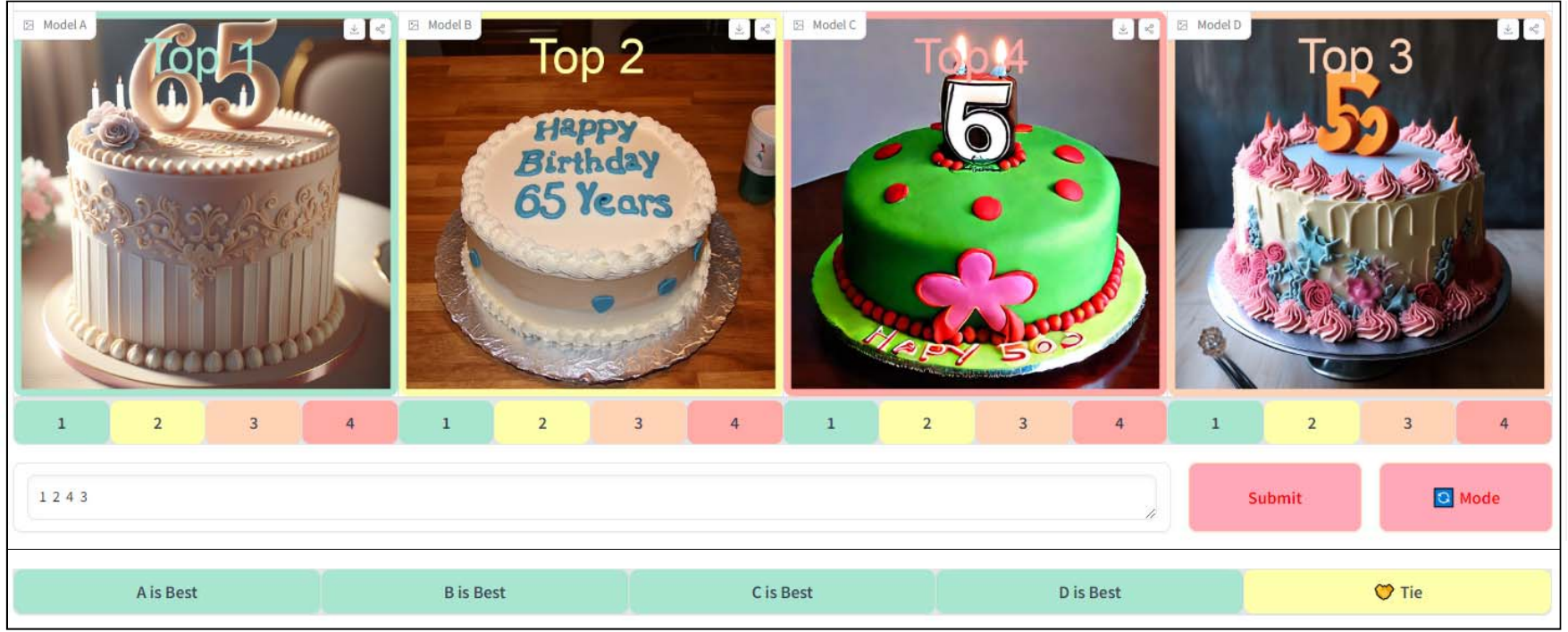}
        \caption{Display of the voting interface.}
    \end{subfigure}
    \caption{Interface of K-Sort Arena served by Huggingface Space.}
    \label{fig:layout}
\end{figure*}

\section{Interface Layout}
\label{app:layout}
K-Sort Arena is served by Huggingface Space, and we carefully design the interface based on gradio to achieve a proper layout and user-friendly interaction.
The interface layout is shown in Figure \ref{fig:layout}.
First, we describe the initial interface before model running, which is divided into three main regions.
\begin{itemize}
    \item Region \textcircled{1} describes the background of the project and the evaluation rules, and serves as a guide for users to vote.
    \item Region \textcircled{2} is the prompt input window, which allows users to enter their own prompts or click ``Random Prompt" to randomly select from the data pool.
    \item Region \textcircled{3} is some completed samples, including the prompt-image sample pairs, which allow users to quickly complete an experience without running the model.
\end{itemize}

After finishing the model running, the interface automatically jumps to the voting interface. It supports two voting modes, and users can click ``Mode" to switch between them.
\begin{itemize}
    \item In Rank Mode, there are 4 buttons below each image to indicate its rank. Whenever a user clicks on it, the image is retouched with responsive borders and markup.
    \item In Best Mode, users can choose the best model or a tie.
\end{itemize}

\section{User Behavior Analysis}

We conduct a comprehensive analysis of user effort in a visual voting task by collecting behavioral data from ten trained participants, and the results are shown is Figure \ref{fig:rebuttal_1}. Our study examines effort expenditure across different values of K and varying levels of prompt complexity. Notably, we observe that the additional effort required for K=4 compared to K=2 remains within an acceptable range due to the perceptual intuitiveness of the task. This suggests that while increasing K introduces more choices, the cognitive load does not escalate significantly, allowing users to make selections with relative ease.

Furthermore, as prompt complexity increases, particularly with long prompts derived from the DiffusionDB dataset, users naturally spend more time reading and processing the information. This extended reading phase effectively diminishes the relative differences in effort when engaging in visual voting, as the majority of cognitive load is shifted towards comprehension rather than selection. 

\end{document}